\documentclass[10pt,twocolumn,letterpaper]{article}

\usepackage[pagenumbers]{cvpr}

\usepackage{graphicx}
\usepackage{amsmath}
\usepackage{amssymb}
\usepackage{booktabs}

\usepackage{hyperref}
\hypersetup{colorlinks,breaklinks}

\def\cvprPaperID{*****}
\def\confName{CVPR}
\def\confYear{2022}

\newcommand{\xhdr}[1]{\vspace{2pt}\noindent\textbf{#1}}

\newcommand{\figref}[1]{Fig.~\ref{#1}\xspace}
\newcommand{\secref}[1]{Sec.~\ref{#1}\xspace}

\usepackage[capitalize]{cleveref}
\crefname{section}{Sec.}{Secs.}
\Crefname{section}{Section}{Sections}
\Crefname{table}{Table}{Tables}
\crefname{table}{Tab.}{Tabs.}

\begin{document}

\title{Instance-Specific Image Goal Navigation:\\Training Embodied Agents to Find Object Instances}

\author{
    \textbf{Jacob Krantz$^{1}$\thanks{Work done while interning at Meta AI's FAIR Labs.\newline Correspondence: \href{mailto:krantzja@oregonstate.edu}{krantzja@oregonstate.edu}}}
    ~~~\textbf{Stefan Lee$^{1}$}
    ~~~\textbf{Jitendra Malik$^{2,3}$}
    ~~~\textbf{Dhruv Batra$^{2,4}$}
    ~~~\textbf{Devendra Singh Chaplot$^{2}$} \\
    $^{1}$Oregon State University~~$^{2}$Meta AI~~$^{3}$UC Berkeley~~$^{4}$Georgia Tech
}
\maketitle

\begin{abstract}
   We consider the problem of embodied visual navigation given an image-goal (ImageNav) where an agent is initialized in an unfamiliar environment and tasked with navigating to a location `described’ by an image. Unlike related navigation tasks, ImageNav does not have a standardized task definition which makes comparison across methods difficult. Further, existing formulations have two problematic properties; (1) image-goals are sampled from random locations which can lead to ambiguity (\eg looking at walls), and (2) image-goals match the camera specification and embodiment of the agent; this rigidity is limiting when considering user-driven downstream applications. We present the Instance-specific ImageNav task (InstanceImageNav) to address these limitations. Specifically, the goal image is `focused’ on some particular object instance in the scene and is taken with camera parameters independent of the agent. We instantiate InstanceImageNav in the Habitat Simulator using scenes from the Habitat-Matterport3D dataset (HM3D) and release a standardized benchmark to measure community progress.
\end{abstract}

\section{Introduction}
\label{sec:intro}

\begin{figure}[t]
    \centering
    \includegraphics[width=\columnwidth]{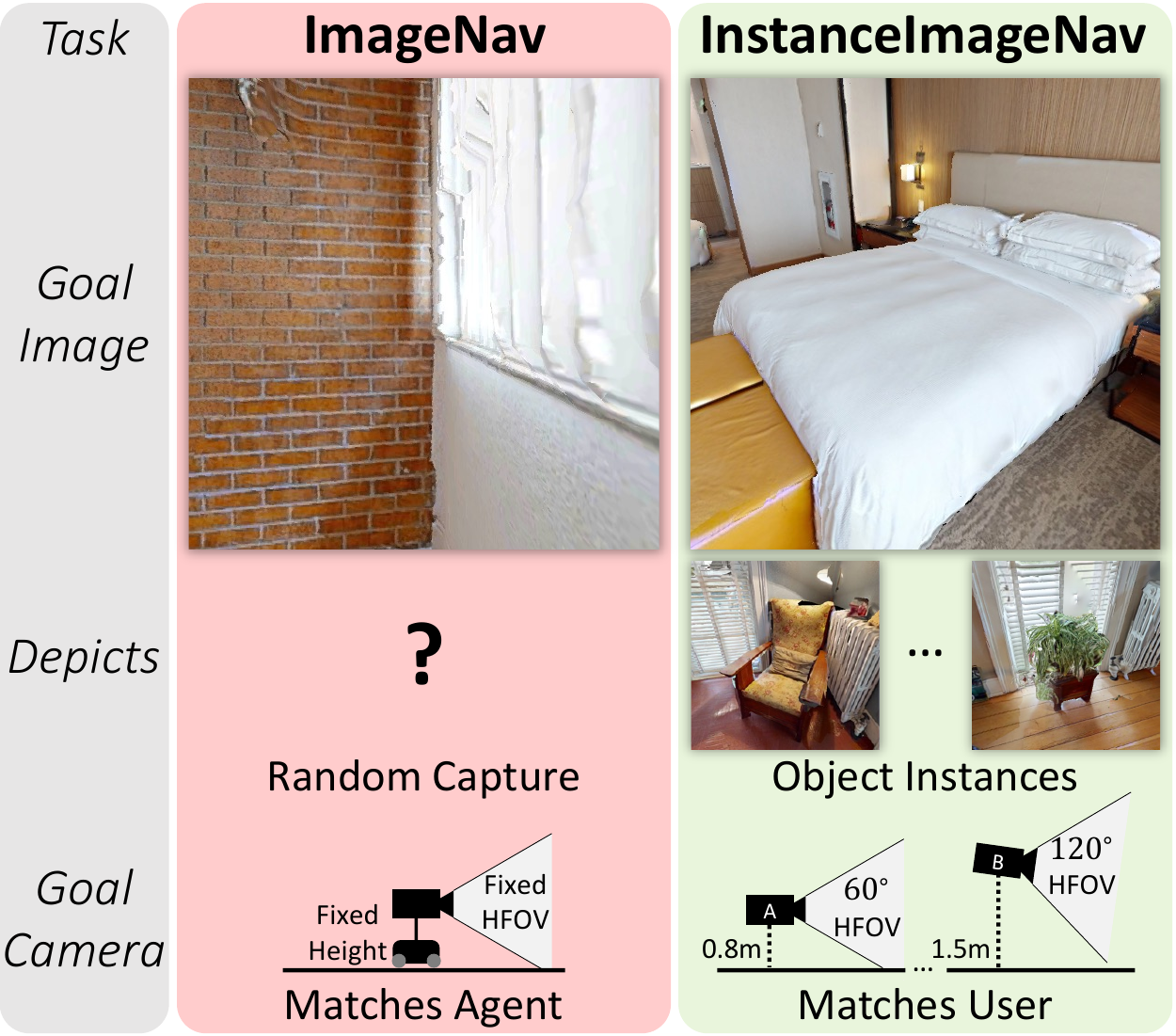}
    \caption{We present InstanceImageNav where an agent is tasked with navigating to the object depicted by a goal image. The goal camera is reflective of the task issuer, not the task executor.}
    \label{fig:teaser}
\end{figure}

Visual navigation is a prerequisite to creating mobile robot assistants that can aid humans in a variety of tasks at home and at work. Although navigation has been studied in the research community for over three decades~\cite{desouza2002vision}, there has been increased interest in building visual navigation systems in recent years~\cite{zhu2017target, savva2019habitat} following advances in computer vision and machine learning~\cite{he2016deep,kober2013reinforcement,schulman2017proximal}. Many navigation tasks have been proposed corresponding to how navigation goals are specified --- relative coordinates in PointNav, object categories in ObjectNav, or language instructions in vision-and-language navigation (VLN~\cite{anderson2018evaluation}). Most common navigation tasks are standardized and benchmarked across multiple public challenges and leaderboards such as PointNav in Habitat~\cite{habitat2020sim2real}, ObjectNav in Habitat~\cite{habitatchallenge2022} and AI2Thor~\cite{RoboTHOR}, and Vision-and-Language Navigation in Matterport3D~\cite{Anderson_2018_CVPR,ku2020room} and Habitat~\cite{krantz2020beyond,deitke2022retrospectives}. However, despite a flurry of papers on the topic \cite{zhu2017target,chaplot2020neural,choi2021image,hahn2021no,majumdar2022zson,yadav2022offline}, the ImageNav task still lacks a standardized task definition, dataset, and benchmark. Consequently, the range of prior ImageNav works often uses inconsistent task definitions, sensor specifications, agent embodiments, and different (and sometimes non-public) training and evaluation datasets, making it difficult to compare different methods.

This lack of standardization is partly due to the generality of image-based goal specification. At a high level, the ImageNav task involves an agent initialized in an unfamiliar environment and tasked with navigating to a goal location that is `described' to the agent in terms of an image. But what precisely do we mean by `described in terms of an image'? Prior work has operationalized this to mean `an image taken at a random goal location using precisely the same sensor specification and embodiment as the agent'. While this is a reasonable choice, a randomly-sampled location and viewpoint often results in an image depicting a non-discriminative region in the environment (\eg a wall or hallway) and simply may not contain enough information to unambiguously identify the goal location. Chaplot et al.~\cite{chaplot2020neural} ameliorate this issue by extending the agent's field of view to a full panoramic 360$^{\circ}$. However, this formulation limits the number of robot platforms a learned policy can be transferred to, as commonly-deployed RGBD sensors have a narrow field of view. Must the goal-image specification match the camera specification of the agent?

We argue that sampling image-goals at random locations and tying the imaging of goal locations to agent embodiments are not representative of real-world applications. The most natural application of ImageNav is where the user wants to specify a particular object instance as the goal. For instance, if the user wants a robot assistant to check whether a laptop is on \textit{their} desk, they can ask the robot to find their desk by showing an image of their desk. It can be severely limiting to insist that the user use the identical camera as the one mounted on the robot and match the pose of the camera to one attainable by the robot (\eg, 3 feet from the ground). These application-guided considerations were not captured in prior ImageNav work.

In this paper, we address these limitations and present the task of Instance-specific ImageNav (InstanceImageNav; \figref{fig:teaser}). InstanceImageNav follows the same spirit as prior work --- a navigation task where goals are specified via an image. However, it makes subtle but important design decisions that overcome the limitations of prior work detailed above. Specifically, (1) the goal image is not expected to match the sensor specification or embodiment of the navigating agent, and (2) the goal image is `focused' on an object instance rather than randomly sampled in the scene. The benefit of (1) is increased application realism as users might capture goal images using a different camera and pose than the agent, and increased generality as models need not be retrained in such an event. The benefit of (2) is decreased task ambiguity as image goals depict objects contextualized in the scene, and further application realism as navigating to object instances is a prerequisite to solving scene understanding and object manipulation tasks. We instantiate InstanceImageNav in the Habitat Simulator \cite{savva2019habitat} using Habitat-Matterport3D (HM3D) scenes~\cite{hm3d} annotated with HM3DSEM semantics~\cite{hm3d-semantics}. We provide an episode dataset and will release a public leaderboard on EvalAI to standardize evaluation and spur progress.

\section{InstanceImageNav Task Definition}
\label{sec:task}

\begin{figure*}[t]
    \centering
    \includegraphics[width=0.9\textwidth]{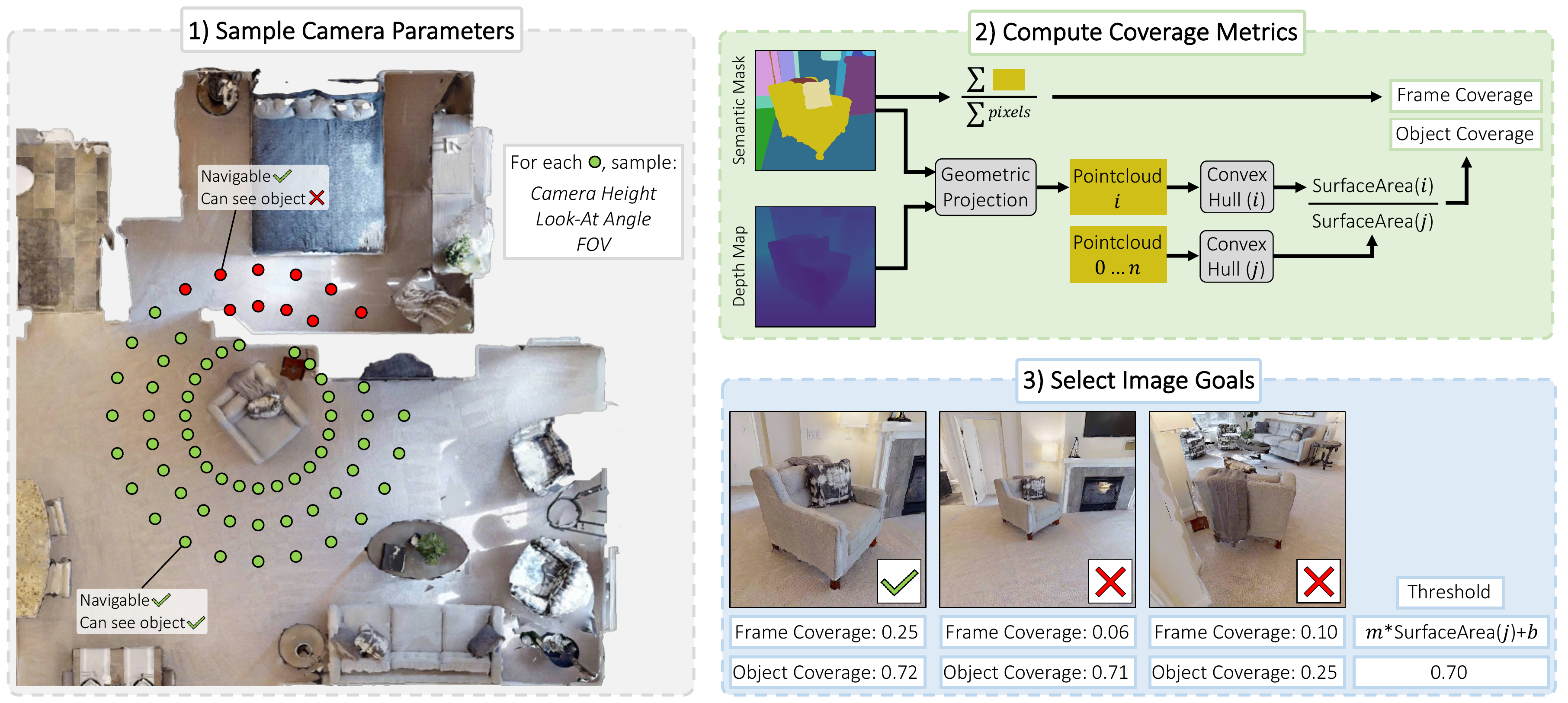}
    \caption{Goal image generation: In this example, the target is an armchair. We sample candidate camera parameters radially about the object. For each candidate's RGBD+Semantic render, we compute object coverage (how much of the object is seen) and frame coverage (how much of the image is the object). We threshold these values to select goal images with clear and natural views of the target object.}
    \label{fig:goal_heuristic}
\end{figure*}

The InstanceImageNav task is defined as the task of navigating to a specific object instance in an unexplored environment specified by an RGB image that prominently depicts that object. The agent is initialized at a random pose in the environment and must find a specific object instance given egocentric vision and the image goal indicating the target object.

Notably, this task differs from the ObjectNav task, not only in how the goal is specified to the agent (RGB image \vs an object category label), but also in that the goal directs the agent to a \emph{specific} object instance. If the image goal depicts a bed, the agent must navigate to that \textit{specific} bed — stopping near any other bed (admissible behavior in ObjectNav) constitutes a failure to solve the InstanceImageNav task.

Grounding this task in a fully-specified benchmark can facilitate precise experimental comparisons and a better calibrated notion of community progress. An InstanceImageNav benchmark ought to precisely define the following:

\xhdr{Goal Image Criteria:} What characteristics must an image satisfy to be considered a valid depiction of a goal object?

\xhdr{Evaluation Protocol:} What constitutes a successful navigation? Should the agent navigate to the object or to the precise location the image was taken?

\xhdr{Agent Embodiment:} What is the observation space, the action space, and the bodily system of the agent?

\xhdr{Environments:} What scenes should be used? What object instances are considered valid navigation goals?

We detail a system of answers to the above questions and in \secref{sec:hm3d_benchmark} provide a concrete instantiation in the Habitat Simulator \cite{savva2019habitat} with scenes from the Habitat-Matterport 3D Dataset (HM3D) \cite{hm3d,hm3d-semantics}.

\subsection{Goal Image Criteria}

What behavior can be achieved via image-guided navigation? We posit that images can serve as a general-purpose descriptor of object instances. For example, an image of a couch in one room clearly disambiguates a reference to that couch \vs a couch in any other room. Thus, it is natural to connect an image to the act of navigating to the depicted object instance. This skill is relevant to achieving higher-level semantic goals like \textit{``check if the laptop is on my desk"}. Here, we detail recommended criteria for what defines a valid goal image.

\xhdr{Depicts an object instance unambiguously.} An object's image goal must represent the object in some manner. While this can be achieved abstractly (\eg, a sketch~\cite{al2022zero}), we recommend direct visual capture. However, not all images containing an object represent it clearly; the camera could be too close or too far, other objects could be equally (or better) represented, or an object has a generic appearance requiring in-situ context to uniquely identify, such as a toilet. To mitigate possible ambiguity, clear image goals will maximize single-object representation and minimize multi-object ambiguity.

\xhdr{Independent of agent embodiment.} Goal images should reflect camera parameters and poses of the entity issuing the task, not of the agent executing the task. This reflects the context of a realistic deployment of an ImageNav model where images could be captured from arbitrary heights and with cameras of varying focal lengths. Sampling goal images from a distribution of these values ensures models are evaluated against this criteria.

\subsection{Evaluation Protocol}

Previous ImageNav tasks define successful navigation as reaching the location from which an image was captured \cite{chaplot2020neural,hahn2021no}. However in InstanceImageNav, image-goals refer to an instance in the scene. Such a reference can be made from many different camera locations (\eg, the green candidate locations in \figref{fig:goal_heuristic}). Accordingly, an agent should navigate to the referenced object rather than the source location of the image. Batra et al.~\cite{batra2020objectnav} detail notions of success and efficiency in navigating to an object instance when recommending an evaluation protocol for the ObjectNav task. Success conditions included intentionality (agent emits a STOP action), validity (the agent’s pose is possible relative to its embodiment), proximity (the agent is within some distance threshold of the object), and oracle visibility (the object can be seen from the agent’s position by looking around). While this criteria was generalized to a set of valid object goals for ObjectNav, it can be applied to the single instance depicted by the image-goal in InstanceImageNav, as we do in \figref{fig:goal_heuristic}.

\subsection{Agent Embodiment}

Embodiment can largely be decided in the more general context of visual navigation where decisions focus on visual decision-making, speed of simulation, and future transfer to the physical world (Sim2Real) \cite{habitat2020sim2real,batra2020objectnav}. Beyond this, matching embodiment to other tasks (\eg ObjectNav) can better enable the study of multi-task/transfer agents and unify Sim2Real experiments. Additionally, an InstanceImageNav agent must have a sensor suite that affords recognizing the object depicted by an image goal, such as a color sensor.

\subsection{Environment}

Environments should contain a large set of objects from which to generate image goals and evaluate an agent's instance-level object recognition ability. While an image goal could depict any recognizable object, we recommend starting with a small, closed-world semantic vocabulary of common objects to help focus the task on the visual recognition and disambiguation of \textit{instances} within \textit{categories}. Vocabulary expansion and open-world vocabulary~\cite{al2022zero} can be future extensions.

\section{The InstanceImageNav-HM3D Benchmark}
\label{sec:hm3d_benchmark}

\begin{figure*}[t]
    \centering
    \includegraphics[width=0.9\textwidth]{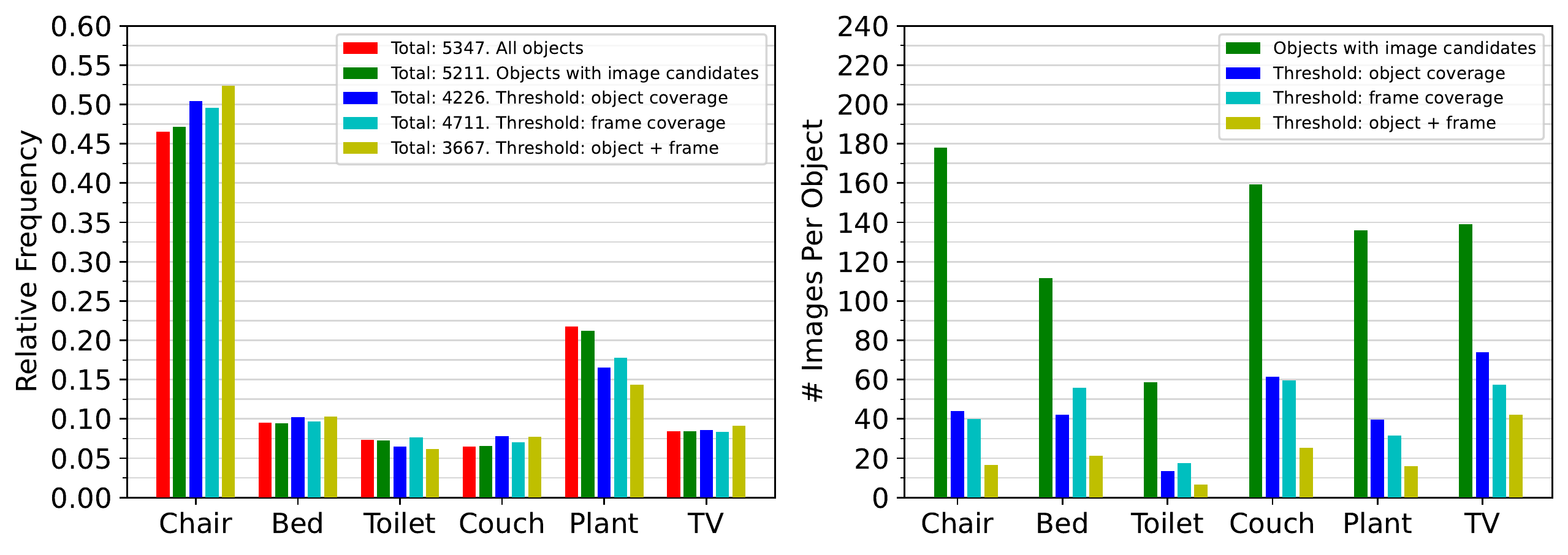}
    \caption{Distributions of object categories and image goals in the InstanceImageNav-HM3D Train split. Left: distribution of objects at each stage of image filtering. Right: average number of image goals per object instance at each stage of image filtering.}
    \label{fig:obj_distributions}
\end{figure*}

\begin{figure*}[t]
    \centering
    \includegraphics[width=0.9\textwidth]{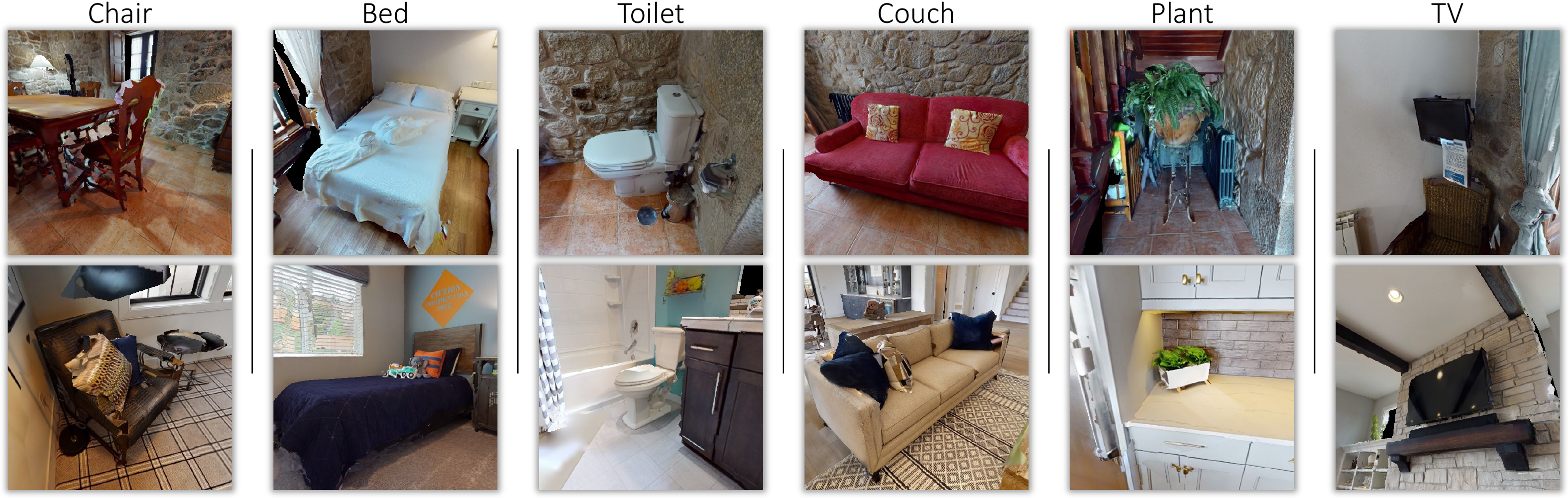}
    \caption{Example goal images in InstanceImageNav-HM3D for each object category. Reflecting how the task may be issued in the real world, these images display a wide diversity of capture settings, including variable camera heights, field of view, and distance to the object.}
    \label{fig:goal_diversity}
\end{figure*}

We instantiate InstanceImageNav in the Habitat Simulator using the Habitat-Matterport3D Dataset (HM3D) with its semantic annotations. This benchmark is publicly available\footnote{\href{https://github.com/facebookresearch/habitat-lab}{github.com/facebookresearch/habitat-lab}} and a leaderboard will be established. We detail the implementation of this benchmark and dataset below.

\xhdr{Scenes.} This benchmark is situated in the indoor, photorealistic scenes of HM3D using the semantically-annotated HM3DSEM subset~\cite{hm3d-semantics} (split 145 for train, 36 for validation, and 35 for test). HM3D was selected because it is the largest publicly-available dataset of photorealistic, semantically-annotated indoor spaces.

\xhdr{Objects.}
Consistent with literature in ObjectNav \cite{chaplot2020object, habitatchallenge2022}, we include all object instances belonging to the following six semantic categories: `\textit{Chair}', `\textit{Bed}', `\textit{Toilet}', `\textit{Couch}', `\textit{Plant}', and `\textit{TV}'. In the training split alone, this amounts to 5,347 total objects (\figref{fig:obj_distributions}).

\begin{figure*}[t]
    \centering
    \includegraphics[width=0.9\textwidth]{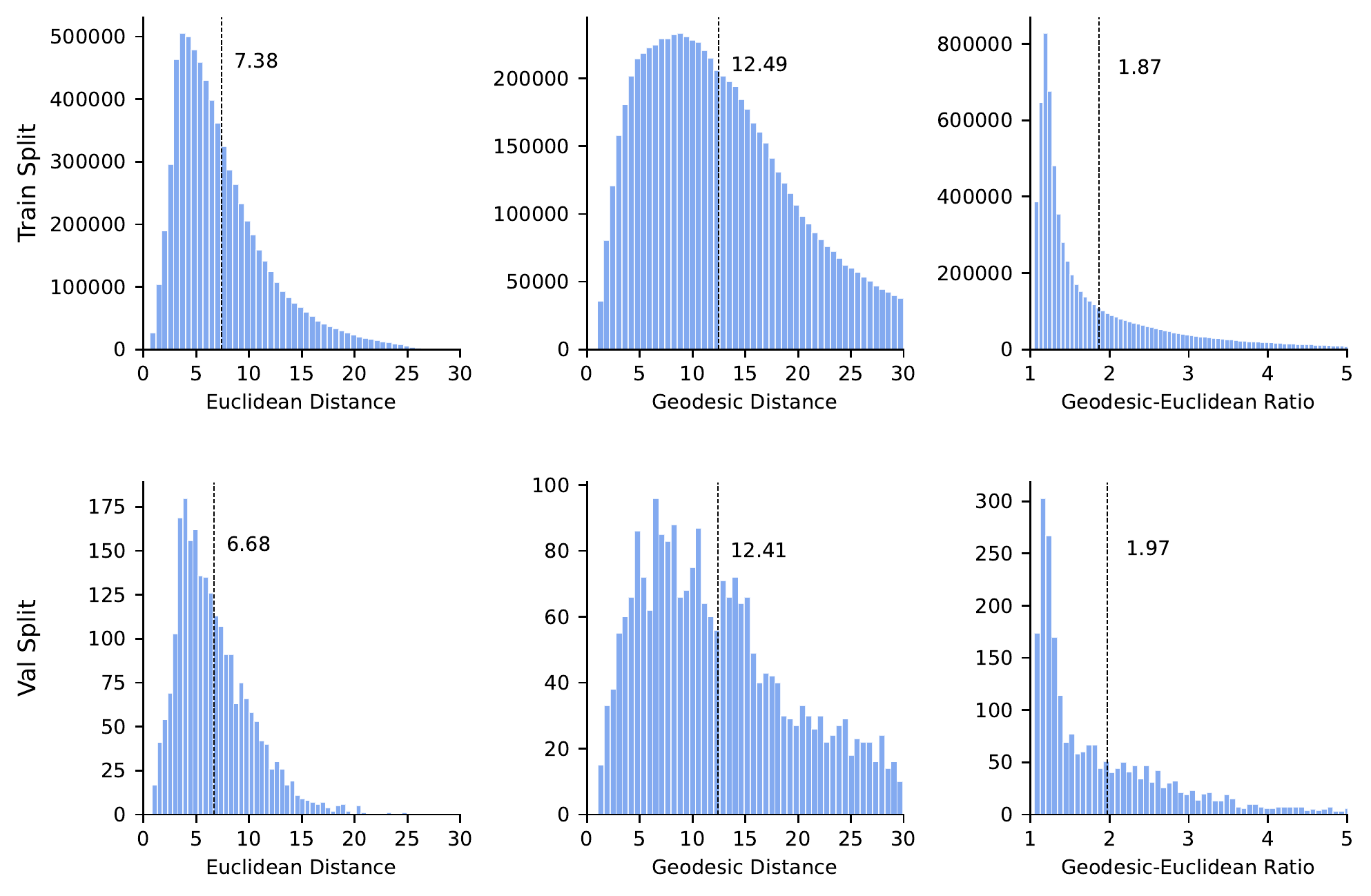}
    \caption{Shortest path statistics for the InstanceImageNav-HM3D dataset. Episodes from the training split (top) and the validation split (bottom) are compared along the axes of Euclidean distance from start to goal (left), the geodesic distance along the shortest path (center), and the ratio of geodesic distance to Euclidean distance (right). Euclidean and geodesic distances are in meters.}
    \label{fig:path_lengths}
\end{figure*}

\xhdr{Image Goals.}
We generate an object's image goals by sampling a set of candidates (\figref{fig:goal_heuristic}(1)), computing coverage heuristics (\figref{fig:goal_heuristic}(2)), and thresholding them (\figref{fig:goal_heuristic}(3)).

We sample candidate images from a calibrated pinhole camera. The camera pose is determined from polar coordinates $(r, \theta)$ relative to the object centroid and a height $h$ where $r \in \left \{ 0.5\text{m}, 1.0\text{m}, 1.5\text{m}, 2.0\text{m} \right \}$, $\theta \in \left \{0^{\circ}, 10^{\circ}, \dots, 360^{\circ} \right \}$, and $h \sim \mathcal{U}(0.8\text{m}, 1.5\text{m})$. The direction of the camera is set such that the camera principle axis is parallel to the ray joining the center of the camera to the centroid of object. We vary the look-at angle by sampling pitch and yaw deltas from $\mathcal{U}(-5^{\circ},5^{\circ})$. The horizontal field of view is sampled from $\mathcal{U}(60^{\circ}, 120^{\circ})$.

We compute coverage heuristics for each resulting candidate image. As shown in \figref{fig:goal_heuristic} (2), we establish two criteria: \textit{frame coverage} (how much of the image is the object) and \textit{object coverage} (how much of the object is seen):

\begin{itemize}
    \item \textit{Frame coverage} ($c_f$) is the ratio of pixels that depict the object to the total number of pixels as computed from a 512x512 semantic mask. High frame coverage ensures the camera is not too far away from the object.
    \item \textit{Object coverage} ($c_o$) is the ratio of the observed object surface area to the object's total observable surface area. We compute observed surface area by projecting the semantic mask into a point cloud and computing the surface area of the resulting convex hull. Similarly, we determine the total observable surface area by projecting all points from all candidate images collectively. High object coverage leads to minimal object ambiguity and enables recognition from novel views.
\end{itemize}

We establish thresholds on $c_o$ and $c_f$ to cull valid image goals from the set of candidates:
\begin{align}
    c_o &> 0.7           &   c_f &> m * \text{OSA} + b
\end{align}
where OSA is the observable surface area of the object and $(m{=}0.0232, b{=}0.02)$ are linear parameters determined to maintain a natural distribution of object categories (\figref{fig:obj_distributions}(left)). The effect of the $c_f$ threshold is that smaller objects require a smaller frame coverage, such as potted plants. After filtering objects without valid image goals, 3,667 objects (69\%) remain in the training split. \figref{fig:obj_distributions} (right) shows the number of image goals before and after coverage thresholds. For the average instance of `\textit{Chair}', 17 representative image goals were selected from 178 possible candidates. \figref{fig:goal_diversity} shows example image goals for each object category.

\xhdr{Agent Embodiment.}
We model agent embodiment after Stretch from Hello Robot\footnote{\href{https://hello-robot.com/stretch-2}{hello-robot.com/stretch-2}}. Specifically, the agent is modeled as a rigid-body, $0^{\circ}$ turn radius cylinder of height 1.41m and radius 0.17m. The agent observes the environment through a forward-facing RGBD camera mounted at 1.31m with a 480x640 resolution and a 58$^{\circ}$ horizontal field of view. The agent is provided GPS+Compass in simulation, which can be operationalized in the real world via the robot operating system (ROS~\cite{ros}). The action space is discrete and consists of \texttt{MOVE\_FORWARD}, \texttt{TURN\_LEFT}, \texttt{TURN\_RIGHT}, and \texttt{STOP} with a 0.25m forward displacement and 30$^{\circ}$ turn angles. All observations and actions are noiseless in simulation.

\xhdr{Evaluation Protocol.} We evaluate an agent’s navigation performance with metrics for success and path efficiency.

\begin{itemize}
    \item \textit{Success}: an episode is successful if the agent issues the \texttt{STOP} command while less than 0.1m away from the nearest valid viewpoint. We define the set of valid viewpoints analogously to the Habitat ObjectNav Challenge; viewpoints are sampled from an $\frac{R_{\text{agent}}}{2}$ grid within 1m of the object’s bounding box. A viewpoint is valid if the object is visible with oracle visibility.
    
    \item \textit{Path efficiency}: we use success weighted by inverse path length (SPL~\cite{anderson2018evaluation}) adapted to a set of goal points, known as ObjectNav-SPL~\cite{batra2020objectnav}. The shortest path length is determined by computing the distance from episode start to each valid viewpoint and taking the minimum.
\end{itemize}

\xhdr{Episode Dataset.}
We create an episode dataset from the above-specified object instances, their valid image goals, and their valid viewpoints. Each episode contains a starting pose ($x, y, z, \theta$) and intrinsic and extrinsic parameters of the image goal. Additional metadata not provided to the agent includes the object category of the object depicted in the image goal, the geodesic distance along the shortest path, the euclidean distance to the object, and the set of valid viewpoints used for evaluation. To sample starting poses, we uniformly sample positions an agent can stand at in the environment and perform rejection sampling based on the following criteria; the geodesic distance must be finite (\eg a narrow hallway cannot be navigated if the agent does not fit) and the ratio of geodesic distance to euclidean distance must be greater than 1.05 to avoid trivial forward-only episodes. We sample 2,000 valid starting poses for each object instance in the training set and assign an equal number of episodes to each image goal, \eg, if an object instance has 10 image goals, then each image goal appears in 200 episodes. For the validation split, we sample 3 starting poses for each object instance. Altogether, the training split has 7,032K episodes and the validation split has 2,340.

\subsection{Challenge}

\xhdr{Model-Free RL Baseline.}
To serve as a task baseline, we employ an end-to-end sensors-to-action network trained with model-free reinforcement learning.

The network architecture we use is closely related to the baselines used in related PointNav, ObjectNav and ImageNav tasks \cite{savva2019habitat, chaplot2020object, hahn2021no}. At each time step, the network observes agent RGB $(\mathcal{V}_{RGB})$, agent depth $(\mathcal{V}_{D})$, the goal image $(\mathcal{V}_G)$, GPS coordinates $(x,z)$, and heading $(\theta)$. Visual observations are encoded with separate ResNet-18~\cite{he2016deep} encoders and GPS and heading are encoded with 32-dimensional linear layers:
\begin{align}
    \bar{\mathcal{V}}_{RGB} &= \text{ResNet50}(\mathcal{V}_{RGB}) \\
    \bar{\mathcal{V}}_{D} &= \text{ResNet50}(\mathcal{V}_{D}) \\
    \bar{\mathcal{V}}_G &= \text{ResNet50}(\mathcal{V}_G) \\
    \bar{\mathcal{P}} &= \text{Linear}(\left [ x, z \right ]) \\
    \bar{\theta} &= \text{Linear}(
        \left [ \cos(\theta), \sin(\theta) \right ]
    ).
\end{align}
These features are concatenated and encoded with a 2-layer LSTM:
\begin{align}
    \bar{\mathcal{V}} &= \left [
        \bar{\mathcal{V}}_{RGB},
        \bar{\mathcal{V}}_{D},
        \bar{\mathcal{V}_G},
        \bar{\mathcal{P}},
        \bar{\theta}
    \right ] \\
    h^{(1)}_t, c^{(1)}_t &= \text{LSTM}(
        \bar{\mathcal{V}}, h^{(1)}_{t-1}, c^{(1)}_{t-1}
    ) \\
    h^{(2)}_t, c^{(2)}_t &= \text{LSTM}(
        h^{(1)}_t, h^{(2)}_{t-1}, c^{(2)}_{t-1}
    ).
\end{align}
Finally, an action is sampled from a categorical distribution:
\begin{align}
    a \sim \text{Categorical}(\text{Linear}(h^{(2)}_t)).
\end{align}

\begin{figure}[t]
    \centering
    \includegraphics[width=\columnwidth]{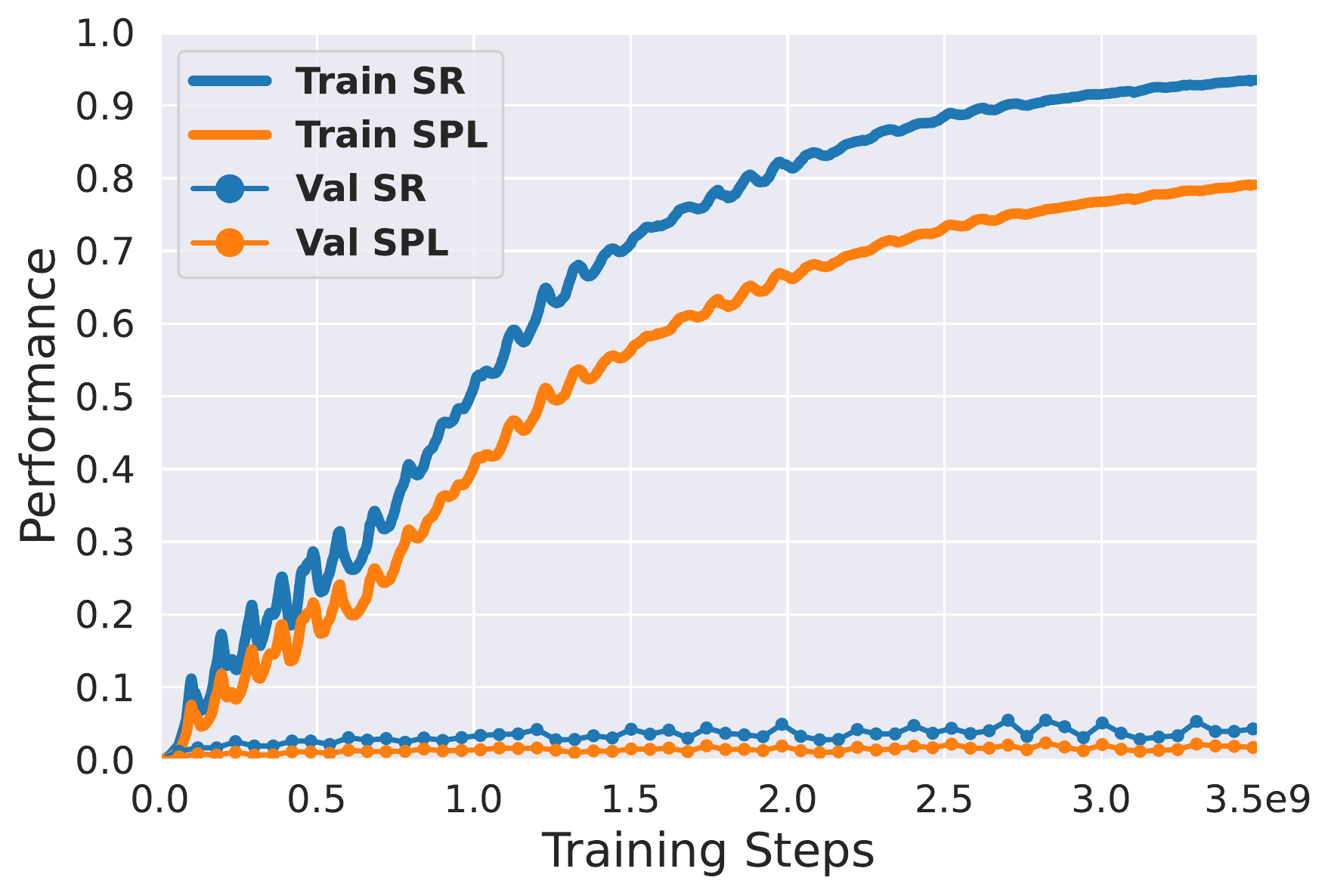}
    \caption{Performance curves for the baseline on training (Train) and validation (Val) as a function of experience steps. The model struggles to generalize with just a 5\% success rate (SR) in Val.}
    \label{fig:baseline_curves}
\end{figure}

This network is trained from scratch using proximal policy optimization (PPO~\cite{schulman2017proximal}) with variable experience rollout (VER~\cite{wijmans2022ver}) distributed across 64 GPUs for 3.5 billion steps of experience. Despite fitting the training set~(\figref{fig:baseline_curves}), this model achieves a success rate of 0.055 and an SPL of 0.023 on validation, indicating poor generalization. Upon release of the leaderboard, this model will be submitted for test split evaluation.

\section{Conclusion}
\label{sec:conclusion}

In this paper, we propose InstanceImageNav, a new benchmark for the ImageNav task that addresses limitations in prior formulations. Instead of randomly sampling ambiguous image goals, we capture images that are well-posed of randomly-sampled objects. Instead of tying together the image goal specification with the agent's sensors and embodiment, we make them independent. We release the InstanceImageNav-HM3D benchmark to standardize evaluation and spur progress in semantic embodied navigation.

{\small
\bibliographystyle{ieee_fullname}
\bibliography{egbib}
}

\end{document}